\documentclass[lettersize,journal]{IEEEtran}
\usepackage{amsmath,amsfonts,amssymb}
\usepackage{algorithmic}
\usepackage{algorithm}
\usepackage{array}
\usepackage[caption=false,font=footnotesize]{subfig} 
\usepackage{textcomp}
\usepackage{stfloats}
\usepackage{url}
\usepackage{verbatim}
\usepackage{graphicx}
\usepackage{cite}
\usepackage{soul}
\usepackage{xcolor}
\usepackage{tabularx}
\usepackage{multirow}
\usepackage{threeparttable}
\usepackage{hyperref}
\usepackage{epstopdf}

\usepackage{enumitem} 
\usepackage{bm}
\usepackage{arydshln}

\newtheorem{rem}{Remark}
\newtheorem{assumption}{Assumption}
\newtheorem{theorem}{Theorem}
\begin{document}


\title{Redundant Observer-Based Tracking Control for Object Extraction Using a Cable Connected UAV}
\author{Benjamin J. Marshall,  
Yunda Yan,
James Knowles, 
Chenguang Yang and Cunjia Liu
\thanks{B.J. Marshall, J.A.C. Knowles, and C. Liu are with the Department of Aeronautical and Automotive Engineering,
 Loughborough University, Loughborough LE11 3TU, U.K. E-mails: \{b.marshall, j.a.c.knowles, c.liu5\}@lboro.ac.uk.}
\thanks{Y. Yan is with the  Department of Computer Science, University College London, London, WC1E 6BT, U.K. E-mail: yunda.yan@ucl.ac.uk.}
\thanks{C. Yang is with Department Of Computer Science, University of Liverpool, Liverpool, L69 3BX, U.K. E-mail: Chenguang.Yang@liverpool.ac.uk.}
}
\maketitle

\begin{abstract}
A new disturbance observer based control scheme is developed for a quadrotor under the concurrent disturbances from a lightweight elastic tether cable and a lumped vertical disturbance. This elastic tether is unusual as it creates a disturbance proportional to the multicopter's translational movement. This paper takes an observer-based approach to estimate the stiffness coefficient of the cable and uses the system model to update the estimates of the external forces, which are then compensated in the control action. Given that the tethered cable force affects both horizontal channels of the quadrotor and is also coupled with the vertical channel, the proposed disturbance observer is constructed to exploit the redundant measurements across all three channels to jointly estimate the cable stiffness and the vertical disturbance. A pseudo-inverse method is used to determine the observer gain functions, such that the estimation of the two quantities is decoupled and stable. Compared to standard disturbance observers which assume nearly constant disturbances, the proposed approach can quickly adjust its total force estimate as the tethered quadrotor changes its position or tautness of the tether. This is applied to two experiments - a tracking performance test where the multicopter moves under a constant tether strain, and an object extraction test. In the second test, the multicopter manipulates a nonlinear mechanism mimicking the extraction of a wedged object. In both cases, the proposed approach shows significant improvement over standard Disturbance Observer and Extended State Observer approaches. A video summary of the experiments can be found in \href{https://youtu.be/9gKr13WTj-k}{this link}.
\end{abstract}

\section{Introduction}
\IEEEPARstart{A}{erial} robots are highly capable platforms combining the manoeuvrability of small unmanned aerial vehicles (UAV) and the dexterity of robotic manipulators. This can enable remote completion of tasks which are valuable to society, ranging from day-to-day activities such as infrastructure cleaning and environmental monitoring, to disaster-response scenarios, through debris removal or object retrieval. Recently, research work has focused on tasks involving direct environmental interaction. This could include object transportation \cite{brandaoSidePullManeuverNovel2022a,villaSurveyLoadTransportation2020}; cleaning  \cite{sunSwitchableUnmannedAerial2021}; mechanism manipulation \cite{marshallNovelDisturbanceDevice2023,byunHybridControllerEnhancing2023}; or power-delivery \cite{jainTetheredPowerSeries2022}. To complete these tasks, UAVs, which are typically multi-copters, need a physical connection with an object. This could include actively controlled manipulators (these may be rigid, multi-link, or bio-inspired \cite{ruggieroAerialManipulationLiterature2018a}) which can complete dexterous and forceful interaction tasks. However, these manipulators increase system weight, and complexity. By using tethers, cables or bungees, forceful interaction can still be achieved at the expense of dexterity \cite{tognonDynamicsControlEstimation2017}.


UAVs using tethers are effective for tasks which involve operating around a fixed point on the ground, or exerting a large force on a single point. Examples of this could include dragging a heavy payload \cite{brandaoSidePullManeuverNovel2022a}, firefighting \cite{droneGuidedHosePipe}, mechanism manipulation \cite{marshallNovelDisturbanceDevice2023}, or for mission-extending power-delivery \cite{jainTetheredPowerSeries2022}. Control objectives in these applications may include trajectory tracking \cite{tognonObserverBasedControlPosition2016}, maintaining stability and safety \cite{yanSurvivingDisturbancesPredictive2023}, winch tension control \cite{nicotraNonlinearControlTethered2017}, transient performance recovery \cite{byunStabilityRobustnessAnalysis2021b,byunHybridControllerEnhancing2023}, minimising payload swing \cite{liangAntiswingControlAerial2022} or improving hover performance by use of the tether forces \cite{sandinoFirstExperimentalResults2016}. In each case, the cable tension imparts a large disturbance on the aerial vehicle, which needs to be modelled and rejected by the flight control design.

\par The disturbance force provided to the UAV from the cable takes two key forms. When the cable is slack, the distributed mass of the cable can be described by the catenary curve model. Indeed, several works use this method for state estimation of the UAV \cite{limaMultimodelFrameworkTetherbased2023a}, state estimation of the cable itself \cite{martinsTensionEstimationLocalization2023} and cooperative UAV-UGV experiments \cite{papachristosEfficientForceExertion2014}. However, once the cable becomes taut, the form of the disturbance changes to that of a spring. This transition between tethered and untethered dynamics is typically dealt with using controllers with multiple operating modes, which can be seen in works such as \cite{brandaoSidePullManeuverNovel2022a,byunHybridControllerEnhancing2023}. During the transition, a high cable stiffness can result in a harsh transition between slack and taut conditions, which in turn creates large transient forces acting on the UAV airframe and its payload. This is often seen in tethered tasks with a moving anchor such as: \cite{brandaoSidePullManeuverNovel2022a,kouraniBidirectionalManipulationBuoy2021}. These difficulties can be reduced by using a compliant cable, which provides two key benefits. First, the extensibility of an elastic cable ``softens" the transition between slack and taut conditions, by lowering the transient force (and therefore acceleration) imparted on the UAV \cite{goodmanGeometricControlLoad2023}. Secondly, since a compliant cable measurably extends compared against a rigid cable, it is possible to estimate and reject the cable forces only using existing onboard sensors, which could lead to reduced vehicle weight and increased endurance. A method to perform this estimation task will be discussed in this paper. To control a multirotor attached to a compliant cable, a simple control law could include a feedforward action, if the cable stiffness is exactly known and constant. However, due to mechanical fatigue, repeated use, or different operating conditions, this may change in over time, or only be approximately known. 

To overcome the influence of a taut cable, existing works typically use methods such as feedback linearisation via differential flatness \cite{tognonDynamicsControlEstimation2017}, hybrid control \cite{byunHybridControllerEnhancing2023}, nonlinear constrained control with reference governor \cite{nicotraNonlinearControlTethered2017} and model predictive control for cooperative experiments \cite{liNonlinearModelPredictive2023}. To improve the performance of tethered UAVs in different applications, many of those tracking controller designs employ a disturbance observer \cite{byunHybridControllerEnhancing2023,byunStabilityRobustnessAnalysis2021b,hanControllerDesignDisturbance2022,yanSurvivingDisturbancesPredictive2023,yukselAerialPhysicalInteraction2019,liangActivePhysicalInteraction2023} or in general state estimators (e.g. high-gain observer in \cite{tognonDynamicsControlEstimation2017} and Kalman filter \cite{reisKalmanbasedVelocityfreeTrajectory2023} to estimate the impact induced by the cable. In each of these observer-based approaches, the designs typically directly estimate the force from the cable, no existing designs exploit the structured and state-dependent nature of the cable force. 


The effect of an elastic cable could be considered as a disturbance which is proportional to UAV position. With this in mind, some inspiration can be taken from contact-inspection tasks. In \cite{markovicAdaptiveStiffnessEstimation2021}, impedance control is used with online stiffness estimation to track a contact-force setpoint. To achieve this, an adaptation law is formed which uses the force-tracking error as feedback. This paper also makes explicit the importance of accurate stiffness estimation for the stability of the impedance filter. Although this sounds promising for a tethered application, the use of force as a feedback state differentiates impedance control from this work, which will only use the sensors required for nominal flight. The adaptation law formed also relies on the vehicle being in a near-steady state to estimate the contact stiffness; which may not be suitable for a tethered case.

\par
\begin{figure}[htb]
    \centering
  \subfloat[UAV attached to anchor point. An exercise band is used as an elastic cable.]{\includegraphics[height=35mm]{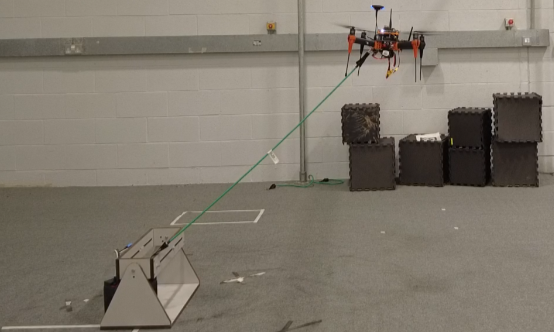}}\label{fig:heavyStand}
    \hfill
  \subfloat[Nonlinear mechanism used to mimic object-extraction task. One side is removed for clarity. Initial position.]{
        \includegraphics[width=40mm]{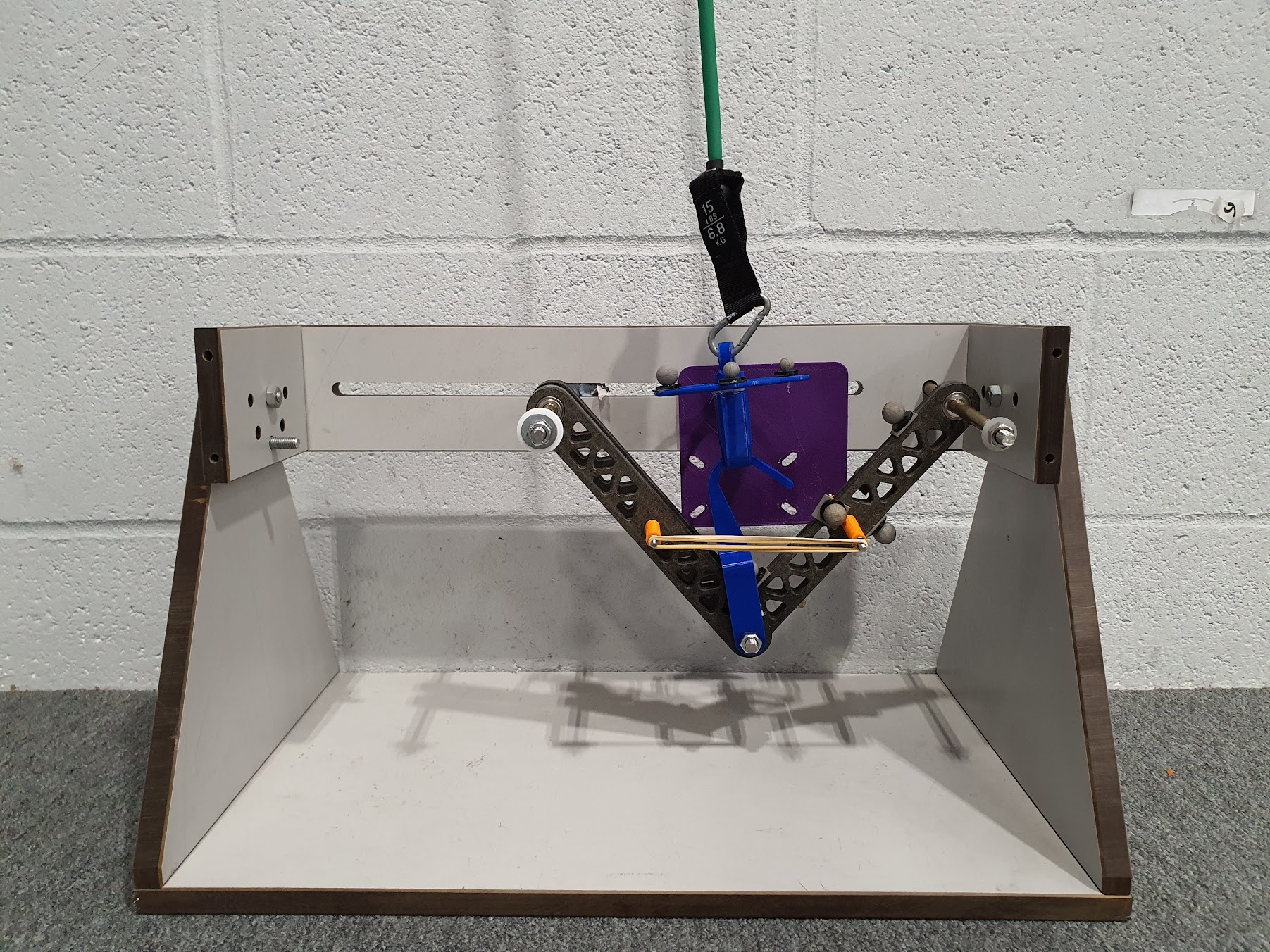}}
       \hfill
       \subfloat[Once force exceeds a critical value, the links move, which releases the drone and hook. Similar to removing an object from the environment.]{\includegraphics[width = 40mm]{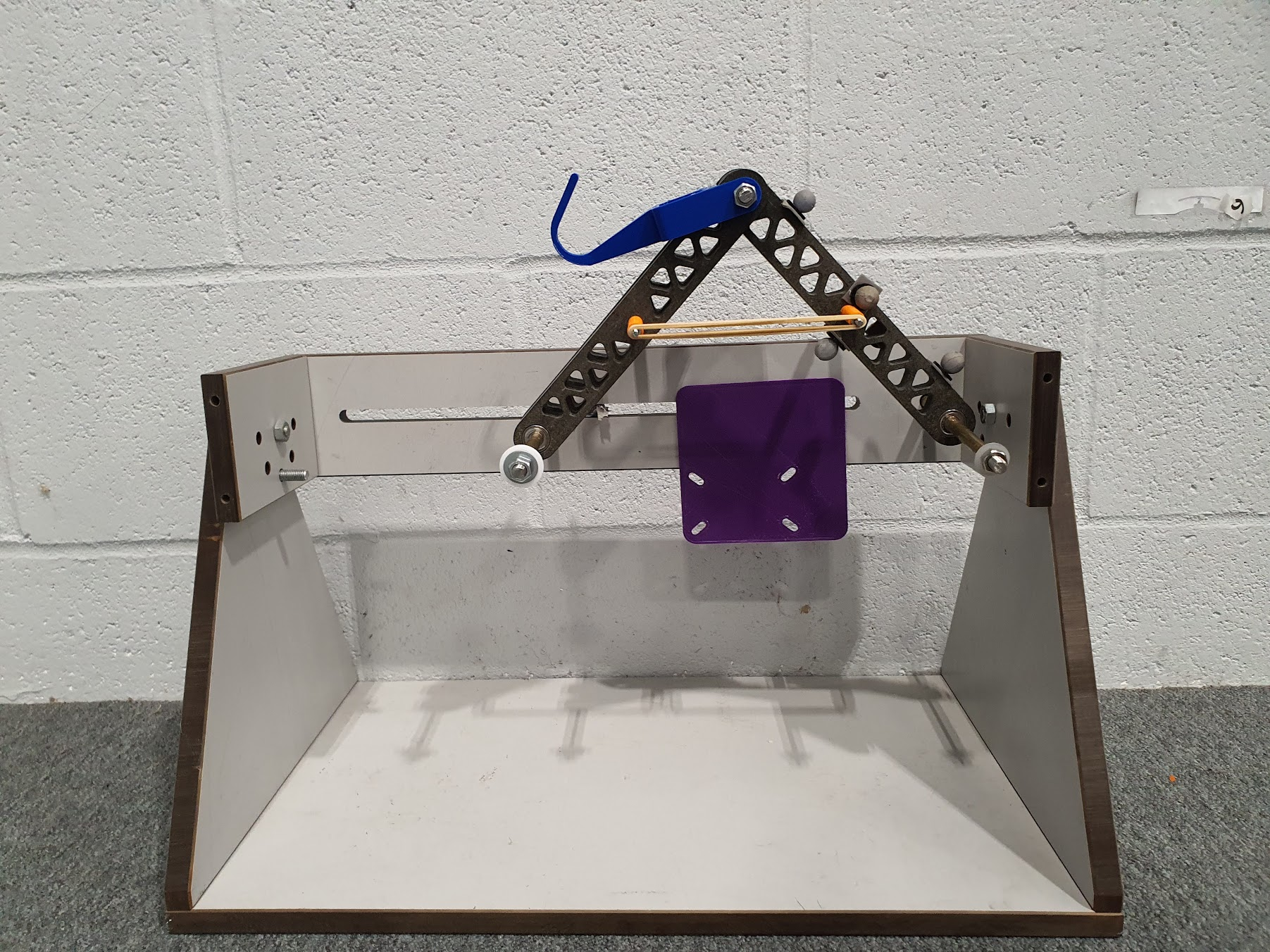}}
  \caption{Experimental setup used for tracking control experiment (top) and object-extraction test (bottom). The object extraction force profile is mimicked using a nonlinear mechanism, which is discussed in\cite{marshallNovelDisturbanceDevice2023}.}
  \label{fig:experimentSetup} 
\end{figure}

 Very recent work often considers interacting with an environment with changing parameters. Such as pushing a cart with a non-constant friction \cite{benziAdaptiveTankbasedControl2022a}, removing a plug from a wall \cite{byunHybridControllerEnhancing2023,byunStabilityRobustnessAnalysis2021b} or moving an articulated object \cite{brunnerPlanningandControlFrameworkAerial2022a}. Tethered UAVs may be able to perform similar tasks, such as forcefully extracting an object from the environment. For example, UAVs could be used to harvest valuable metals out of landfill with an electromagnet-equipped tether, or used to recover important items from underneath rubble during a disaster-response scenario. In this kind of task, the tether will be stretched, then suddenly released as the payload becomes free from the environment. To mimic this kind of disturbance, some studies use electronically triggered releases \cite{byunStabilityRobustnessAnalysis2021b}, mechanical devices \cite{marshallNovelDisturbanceDevice2023}, or manually sever the connection to the environment \cite{rizziRobustSamplingBasedControl2023}. Control methods to overcome these disturbances typically use multiple controller states \cite{byunHybridControllerEnhancing2023}, or limit the power-flow through the dynamic system to enforce safety \cite{cuniatoPowerBasedSafetyLayer2022,benziAdaptiveTankbasedControl2022a}. Each of these designs also employ some kind of force estimation to reject the disturbance from the environment. The former of these two examples \cite{byunHybridControllerEnhancing2023} demonstrates how the overshoot of the overall closed loop system is ultimately bounded based on this hybrid approach. It is notable in this work that the formulated controller requires multiple modes, which complicates the control design and stability analysis. To add to this body of work, this paper will seek to improve the tracking performance of a UAV subject to a disturbance proportional to the vehicle position, as if it were tethered by an elastic tether. The formulated controller will also be demonstrated as effective for object-extraction tasks using the nonlinear mechanism designed in \cite{marshallNovelDisturbanceDevice2023}. Shown in Fig. \ref{fig:experimentSetup} is the tracking performance experiment and nonlinear mechanism that this work will consider.

\par To estimate and reject the state-dependent disturbance from the tether, a new disturbance-Observer based framework is developed in this paper, which allows the estimation task to be expressed as two constants - the cable stiffness coefficient and a lumped uncertainty along the vertical axis. By expressing the estimation problem in this way, the estimation of the total force is updated using the system model. This allows the force estimate to quickly react as the UAV moves. Additionally, the proposed observer can also function in both slack (zero cable stiffness) and taut cable cases. As the UAV used in these experiments concerns the translational motion with three dynamic channels, the position of the UAV is controlled via three virtual inputs. This means there are more inputs and corresponding measurements than the disturbance terms to be estimated, preventing the direct use of any existing Disturbance Observer Based Control (DOBC) design process. This challenge is solved by introducing an novel optimisation process to design observer gain matrix, so that all three channels can be combined and allocated to form a joint estimate for the cable stiffness and the vertical disturbance. In addition to its flexibility, this DOBC has attractive theoretical properties and can be proven to be stable for any operating condition under mild assumptions. The performance evaluation of the new design compared to benchmark approaches is carried out in both simulations and experimental environments. The robustness of the new design in object extraction tasks is also verified in repeatability tests using a specially designed over-centre mechanical device. 

The paper is structured as follows; Sec. \ref{sec:probForms} describes the system dynamics and tracking controller; Sec. \ref{sec:newdesign} designs the Redundant Disturbance Observer (RDO) and Sec. \ref{sec:simResults} evaluates simulation performance. In Sec. \ref{sec:ExpTest} the observer is comprehensively tested. 9 flights are used for each observer to demonstrate experimental consistency. Sec. \ref{sec:concs} presents the conclusions of the paper.


\section{Problem Formulation}\label{sec:probForms}
\par This section details the modelling of a tethered quadrotor, alongside the baseline tracking controller used in the proposed DOBC design. As the majority of the disturbance is applied to the translation dynamics, this work assumes that a low-level controller is available to quickly realise the attitude commands and that it is capable of rejecting the disturbance exerted to the attitude states. By restraining the design to outer-loop control, the proposed DOBC is flexible to different commercially available low-level flight controllers.

\subsection{Dynamic Modelling}
\par
\begin{figure}[h]
    \centering
      \includegraphics[width = 0.4\textwidth]{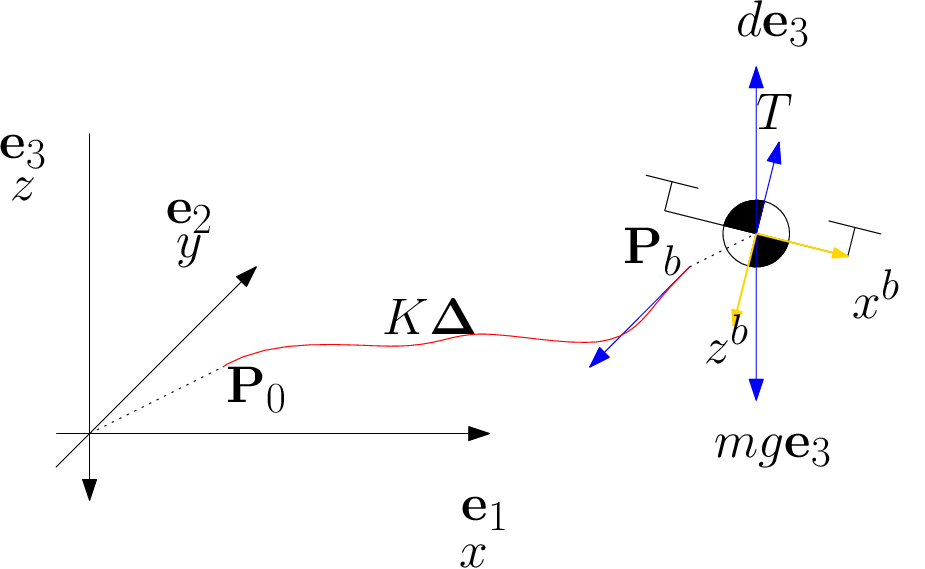}
      \caption{Free body diagram for a quadrotor under the combined disturbances.}
      \label{fig:schematic}
\end{figure}

 Consider a quadcopter of mass $m\in\mathbb{R}_+$, which is at a position $\mathbf{p}_v=\begin{bmatrix} x & y & z \end{bmatrix}^T\in\mathbb{R}^{3\times1}$ and has attitude angles of $\phi,\theta,\psi\in\mathbb{R}$ representing the roll, pitch, and yaw angles respectively. The schematic diagram can be found in Fig.\,\ref{fig:schematic}. The quadcopter is attached to a cable whos end is located at $\mathbf{p}_0\in\mathbb{R}^{3\times1}$. The cable is elastic with a stiffness of $K\in\mathbb{R}_+$ and has an unstrained length of $l_0\in\mathbb{R}_+$. The UAV end of the cable is attached at a location on the quadcopter $\mathbf{p}_b\in \mathbb{R}^{3\times 1}$, defined relative to the centre of gravity. The cable length $l_0$ is known, but the cable stiffness $K$ is to be estimated. The quadcopter moves under a thrust $T\in\mathbb{R}_+$; acceleration due to gravity $g\in\mathbb{R}_+$ and an lumped disturbance force, $d\in\mathbb{R}$. This force $d$ is assumed to act solely along the vertical axis, which captures factors such as an uncertain thrust model, incorrect vehicle mass, ground effects, or the weight of a payload when the attitude angles are reasonably small.

The translation dynamics of the tethered quadrotor are given as:
\begin{equation}\label{drone}
\ddot{\mathbf{p}}_v=\frac{\mathbf{R(\phi,\theta,\psi)}\begin{bmatrix} 0&0&-T \end{bmatrix}^T}{m}+g\mathbf{e}_3+\frac{\mathbf{F}_d}{m}
\end{equation}
where the vertical unit vector is given by $\mathbf{e}_3 = \begin{bmatrix} 0 & 0 & 1 \end{bmatrix}^T\in\mathbb{R}^{3\times1}$, and the total disturbance is defined by $\mathbf{F}_d=-d\mathbf{e}_3-K\mathbf{\Delta}\in \mathbb{R}^{3\times1}$. The function $\mathbf{R(\phi,\theta,\psi)}\in\mathbb{R}^{3\times3}$ defines the rotation matrix from the body-fixed frame to the inertial frame, which comprises three rotations in the order yaw-pitch-roll: 
\begin{multline}
    \mathbf{R}(\phi,\theta,\psi) = \\\begin{bmatrix}
        \text{c}_{\theta}\text{c}_{\psi} & \text{c}_{\psi}\text{s}_{\theta}\text{s}_{\phi} - \text{c}_{\phi}\text{s}_{\theta} & \text{s}_{\phi}\text{s}_{\theta}+\text{c}_{\phi}\text{c}_{\psi}\text{s}_{\theta} \\
        \text{c}_{\theta}\text{s}_{\psi} & \text{c}_{\phi}\text{c}_{\psi}+\text{s}_{\theta}\text{s}_{\phi}\text{s}_{\psi} & \text{c}_{\phi}\text{s}_{\theta}\text{s}_{\psi} -\text{c}_{\psi}\text{s}_{\phi} \\
        \text{c}_{\phi}\text{c}_{\psi}+\text{s}_{\theta}\text{s}_{\phi}\text{s}_{\psi}    &
-\text{s}_{\theta}       & \text{c}_{\theta}\text{s}_{\phi}                     \\
    \end{bmatrix}
\end{multline}
Note that $\text{c}_*$ and $\text{s}_*$ are abbreviations for the cosine and sine functions respectively. To model the behaviour of the cable, the symbol $\mathbf{\Delta} = \begin{bmatrix}
    \Delta_x & \Delta_y & \Delta_z
\end{bmatrix}^T\in\mathbb{R}^{3\times1}$ represents the extension of the cable as a vector of 3 components acting on each translational channel of the quadrotor dynamics. The value of $\mathbf{\Delta}$ is measurable as:
\begin{equation}\label{eqn:deltaDefinition}
\mathbf{\Delta} =\begin{cases} \left(||\mathbf{p}_b^E||-l_0 \right)\frac{\mathbf{p}_b^E}{||\mathbf{p}_b^E||}& ||\mathbf{p}_b^E||-l_0 >0\\
0 & ||\mathbf{p}_b^E||-l_0 \leq 0
\end{cases}
\end{equation}
where $\mathbf{p}_b^E \in \mathbb{R}^{3\times1}$ is the location of the attachment point on the quadcopter expressed relative to the attachment point on the ground. This is expressed as:
\begin{equation}
\mathbf{p}_b^E = \mathbf{p}_v+\mathbf{R}(\phi,\theta,\psi)\mathbf{p}_b-\mathbf{p}_0\end{equation}
It can be seen that the lumped disturbance $\mathbf{F}_d$ is a function of the quadrotor state subject to the unknown stiffness parameter $K$, which makes it difficult to estimate. 

\par To facilitate the DOBC design, the virtual control inputs $\bm{\nu} = \begin{bmatrix} \nu_x & \nu_y & \nu_z \end{bmatrix}^T \in \mathbb{R}^{3\times1}$ are introduced. Physically, these are target forces for the quadrotor to produce along the inertial axes, which are defined as: 
\begin{equation}\label{defnu}
    \bm{\nu} = \mathbf{R}(\phi,\theta,\psi)\begin{bmatrix} 0&0&-T \end{bmatrix}^T = \begin{bmatrix} (\text{s}_{\phi}\text{s}_{\theta}+\text{c}_{\phi}\text{s}_{\psi}\text{s}_{\theta})T \\
    (\text{c}_{\phi}\text{s}_{\theta}\text{s}_{\psi}-\text{c}_{\psi}\text{s}_{\psi})T\\
    \text{c}_{\theta}\text{s}_{\phi}T\end{bmatrix} 
\end{equation}
This simplifies the control design by cancelling the nonlinear terms in the translation dynamics. By using this transformation, each axis to be considered as to have a separate control input. The virtual inputs are then converted to a thrust-attitude command which is sent to the low-level controller. This command is given by:
\begin{equation}\label{eqn:thrust-att-commands}
\begin{aligned}
\theta_d&=\text{atan}\left((\nu_x\text{c}_{\psi_d}+\nu_y\text{s}_{\psi_d})/\nu_z\right)\\
\phi_d&=\text{atan}\left(c_{\theta_d}(\nu_x\text{s}_{\psi_d}-\nu_y\text{c}_{\psi_d})/(\nu_z)\right)\\T&=\nu_z/(\text{c}_{\theta_d}\text{c}_{\phi_d})
\end{aligned}
\end{equation}
where the setpoint yaw angle $\psi_d$ can be set arbitrarily by the operator. It is assumed that these attitude commands are quickly reachable by the low-level controller, and the influence of the disturbances on the low-level controller are rejected by the existing controller. Hence, the control task is to find $\bm{\nu}$ such that $\mathbf{p}_v$ converges to a reference location $\mathbf{p}_r = \begin{bmatrix}
    r_x & r_y & r_z
\end{bmatrix}^T\in\mathbb{R}^{3\times1}$. To achieve this, a basic tracking controller is designed. This tracking controller will then be improved with a disturbance observer to exploit the structured nature of the concurrent disturbances $K\mathbf{\Delta}$ and $d$. 

\subsection{Controller Architecture and Tracking Control}

\par
\begin{figure}[h]
    \centering
    \includegraphics[width = 0.4\textwidth]{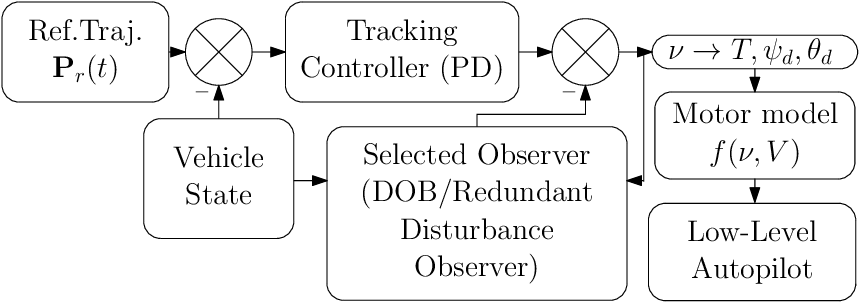}
    \caption{Overall controller structure of PD tracking controller and disturbance observer. The control action $\mathbf{\nu}$ is defined in \eqref{defnu}. $V$ refers to battery voltage.}
    \label{fig:blockDiagram}
\end{figure}

 To follow a reference trajectory in nominal flight, a proportional-derivative (PD) controller is designed first, which will later be augmented by a nonlinear disturbance observer to estimate $K$ and $d$. Following the lumped disturbance model about $\mathbf{F}_{d}$, the disturbance estimate can be reconstructed and compensated via a feed-forward control action. The overall control structure is shown in Fig. \ref{fig:blockDiagram}, system architecture for implementation is discussed later in Sec. \ref{sec:ExpTest}.

\par To design the reference tracking PD controller, the tracking error signal $\mathbf{e}_p$ is defined as
\begin{equation}\mathbf{e}_p = \mathbf{p}_r-\mathbf{p}_v.\end{equation}
The control law is then written as: 
\begin{equation}\label{eqn:controlLaw}
\begin{bmatrix} 
\nu_x & \nu_y & \nu_z 
\end{bmatrix}^T
=m \left({k}_p\mathbf{e}_p+ {k}_d\dot{\mathbf{e}}_p-g\mathbf{e}_3\right)-\hat{\mathbf{F}}_d
\end{equation}
where 
$k_p,k_d\in\mathbb{R}_+$ are proportional and derivative gains, respectively, such that system matrix of the error dynamics 
\begin{equation}
    \mathbf{A}_c=\begin{bmatrix}
        0 & 1\\
        -k_p& -k_d
    \end{bmatrix}
\end{equation}
is Hurwitz. The value $\mathbf{\hat{F}}_d \in \mathbb{R}^{3\times1}$ refers to the total force estimate from the disturbance observer. To form this estimate we aim to exploit the structured nature of $\mathbf{{F}}_d$, this means breaking $\mathbf{F}_d$ into its component parts of $K$ and $d$. To estimate each part of the disturbance force, a disturbance observer will be derived next. 

\section{Redundant Observer Design}\label{sec:newdesign}
 \par This section discusses the design of the Redundant Disturbance Observer (RDO) which is used to form the estimate $\mathbf{\hat{F}}_d$. The goal of this observer is to combine the sensor measurements in three translation directions to estimate the two concurrent disturbance terms. By reducing the estimation task to estimating two constants (the stiffness $K$, and force $d$), the observer can quickly update its total force estimate $\hat{\mathbf{F}}_d$ for a new system state using the function $\mathbf{\Delta}$ defined in \eqref{eqn:deltaDefinition}. This differs from existing methods which may assume the time derivative of the disturbance approximately zero. To begin the design, the following assumptions are necessary:
\begin{assumption}[Sensor settings] \label{assu:Priors}\label{sensor} It is assumed that the following signals can be measured precisely:
\begin{enumerate}[label=(\roman*)] 
    \item System states $\mathbf{p}_v$ and $\dot{\mathbf{p}}_v$ are measurable.
     \item Cable extension $\mathbf{\Delta}$ is measurable based on \eqref{eqn:deltaDefinition} if $\mathbf{p}_0$ is measurable and $l_0$ is constant.
\end{enumerate}
\end{assumption}
\begin{assumption}\label{consdis}
    The cable stiffness $K$ and the vertical force $d$ are unknown constants.
\end{assumption} 

\par Instead of estimating the lumped disturbance $\mathbf{F}_d$ directly as in some existing works (see e.g., \cite{liuTrackingControlSmallscale2012, leeAerialManipulatorPushing2021,reisKalmanbasedVelocityfreeTrajectory2023,yanSurvivingDisturbancesPredictive2023}), the proposed approach estimates the stiffness $K$ and force $d$ simultaneously, such that the final force estimate can be constructed as:
\begin{equation}\label{estFd}
  \hat{\mathbf{F}}_d=-\hat{d}\mathbf{e}_3-\hat{K}\mathbf{\Delta}
\end{equation}
where $\hat{\mathbf{F}}_d$, $\hat{K}$, and $\hat{d}$ are the estimates of ${\mathbf{F}}_d$, ${K}$, and ${d}$.



The first step in constructing the nonlinear disturbance observer is to specify the desired estimation dynamics, such that the estimate of the unknown stiffness $K$ and force $d$ can converge simultaneously. These target estimation dynamics can be written as:
\begin{equation}\label{estdyn}
\begin{bmatrix}
    \dot{\hat{K}} \\\dot{\hat{d}}
\end{bmatrix}= \mathbf{A}_o \begin{bmatrix}
    \hat{K} - K\\\hat{d} - d
\end{bmatrix},~\mathbf{A}_o=-\begin{bmatrix}
    c_1 &0\\
    0&c_2
\end{bmatrix}
\end{equation}
where $c_1,c_2\in\mathbb{R}_+$ are constants. Note that \eqref{estdyn} cannot be implemented due to the direct use of the terms $\hat{K} - K$ and  $\hat{d} - d$. However, by substituting \eqref{defnu} into \eqref{drone} it can be shown that
\begin{equation}\label{diseq}
    d\mathbf{e}_3+K\mathbf{\Delta}=\bm v+mg \mathbf{e}_3 -m\ddot{\mathbf{p}}_v
\end{equation}
which shows that the unknown stiffness $K$ affects the full dynamics through the force $K\mathbf{\Delta}$ and the unknown disturbance force $d$ is applied only in the $z$-direction. Since $\bm{\Delta}$ can be measured, the error term $(\hat{K} - K)\bm{\Delta}$ can be used. As all three translation directions are affected by $K\bm{\Delta}$, which has a single unknown parameter $K$, this introduces the redundancy built-into this design. After estimating $K$, it is then necessary to remove the vertical component, $\hat{K}\Delta_z$ from the vertical component of $\bm{\hat{F}}_d$ to estimate $d$.


Starting with the desired error dynamics in \eqref{eqn:desiredErr}, and invoking assumption \ref{consdis} to obtain \eqref{eqn:invokeA1}, the error dynamics are given by the following:
\begin{subequations}\label{estdyn2}
  \begin{align}
  \begin{bmatrix}
      \dot{\hat{K}} - \dot{\hat{K}}\\\dot{\hat{d}} - \dot{d}
  \end{bmatrix} &= \mathbf{L}(\cdot)\left(\hat{\mathbf{F}}_d - \mathbf{F}_d\right)\label{eqn:desiredErr}\\
     \begin{bmatrix}
\dot{\hat{K}}  \\\dot{\hat{d}} 
\end{bmatrix}&=\mathbf{L}(\cdot)\begin{bmatrix} (\hat{K}-K) \Delta_x \\ (\hat{K}-K) \Delta_y  \\ (\hat{K}-K) \Delta_z +(\hat{d}-d) \end{bmatrix}\label{eqn:invokeA1}\\
&=\begin{bmatrix}
    \mathbf{l}_{1:3}^T\\
    \mathbf{l}_{4:6}^T
\end{bmatrix}\begin{bmatrix}
   \mathbf{\Delta}  &  \mathbf{e}_3   
\end{bmatrix}\begin{bmatrix}
    \hat{K} - K\\\hat{d} - d
\end{bmatrix}\\
&=\begin{bmatrix}
    \mathbf{l}_{1:3}^T\\
    \mathbf{l}_{4:6}^T
\end{bmatrix} \left(
\hat d \mathbf{e}_3+ \hat K\mathbf{\Delta} -\bm v-mg \mathbf{e}_3 +m\ddot{\mathbf{p}}_v
\right) \label{estdyn2c}
  \end{align}  
\end{subequations}
where
$$\mathbf{L(\cdot)}=\begin{bmatrix}
    l_1&l_2&l_3\\
    l_4&l_5&l_6
\end{bmatrix}=\begin{bmatrix}
    \mathbf{l}_{1:3}^T\\
    \mathbf{l}_{4:6}^T
\end{bmatrix}\in\mathbb{R}^{2\times3},~\mathbf{l}_{1:3},\mathbf{l}_{4:6}\in \mathbb{R}^{1\times3}$$ is the observer gain matrix. The elements of $\mathbf{L(\cdot)}$ are scalar functions which may contain $\bm{\nu},\mathbf{p}_v,\dot{\mathbf{p}}_v,\mathbf{\Delta}$. 

The next step of the disturbance observer design requires to find the observer gain. Ideally, enforcing \eqref{estdyn2} to be the target estimation error dynamics \eqref{estdyn} gives:
\begin{equation} \label{gainrelation}
    \begin{bmatrix}
    \mathbf{l}_{1:3}^T\\
    \mathbf{l}_{4:6}^T
\end{bmatrix}\begin{bmatrix}
   \mathbf{\Delta}  &  \mathbf{e}_3   
\end{bmatrix}=\mathbf{A}_o
\end{equation}
which is equivalent to:
\begin{equation} \label{lsolution0}
        l_3=0,~l_6=-c_2 
\end{equation}
and 
\begin{equation}\label{lsolution1}
    \begin{bmatrix} \Delta_x & \Delta_y & 0 & 0 \\ 0 & 0 & \Delta_x & \Delta_y \end{bmatrix} \begin{bmatrix} l_1 \\ l_2\\ l_4 \\l_5\end{bmatrix} = \begin{bmatrix} -c_1 \\ c_2\Delta_z\end{bmatrix}.
\end{equation}
Note that there are more degrees of freedom in choosing the gains $l_1$, $l_2$, $l_4$ and $l_5$ to satisfy \eqref{lsolution1}. So, to find the optimal combination of observer gains for a given desired estimation dynamics, the pseudo-inverse of \eqref{lsolution1} is used to obtain:
\begin{equation}\label{eqn:gainSelection}
    \begin{bmatrix} l_1 \\l_2\\l_4\\l_5\end{bmatrix}= \frac{1}{\Delta_x^2+\Delta_y^2}\begin{bmatrix} -\Delta_xc_1\\-\Delta_yc_1\\\Delta_x\Delta_zc_2\\\Delta_y\Delta_zc_2\end{bmatrix}
\end{equation}
which holds provided $\Delta_x^2+\Delta_y^2$ is nonzero (see remark \ref{rem:singularity}). 

\par To complete the disturbance observer design, internal variables need to be defined to remove reliance on $\ddot{\bm{p}}_v$, since the acceleration term is not measurable. By rearranging \eqref{estdyn2c}, an internal vector can be defined as $
    \bm \xi=\begin{bmatrix}
        \alpha&\beta
    \end{bmatrix}^T\in\mathbb{R}^{2\times1}$ 
where
\begin{equation}\label{alphabeta}
    \begin{aligned}
        \dot \alpha&= \dot{\hat{K}}-m\mathbf{l}_{1:3}^T\ddot {\mathbf{p}}_v,~ \dot \beta= \dot{\hat{d}}-m\mathbf{l}_{4:6}^T\ddot {\mathbf{p}}_v.
    \end{aligned}
\end{equation}
Taking the indefinite integral of both sides of \eqref{alphabeta} with respect to time gives:
\begin{equation}   \label{DOestimate}
{\hat{K}} = {\alpha} +\gamma_\alpha(\mathbf{\Delta},\mathbf{p}_v),~
{\hat{d}} = {\beta} +\gamma_\beta (\mathbf{\Delta},\mathbf{p}_v)  
\end{equation} 
where 
\begin{equation}\label{gamma}  
\gamma_\alpha(\mathbf{\Delta},\mathbf{p}_v)=m\int \mathbf{l}_{1:3}^Td \dot {\mathbf{p}}_v,~
\gamma_\beta(\mathbf{\Delta},\mathbf{p}_v)=m\int \mathbf{l}_{4:6}^Td \dot {\mathbf{p}}_v
\end{equation} 
and the constant caused by the indefinite integral can be regarded as the initial value of $\bm \xi$ and hence be ignored. 

Finally, the disturbance observer dynamics \eqref{estdyn2c} can be rewritten by using the internal vector defined in \eqref{alphabeta} and the observer gain relation in \eqref{gainrelation}, such that 
\begin{subequations}  \label{DOdyn}
\begin{align}
       \dot{\alpha} & = -c_1\left(\alpha+\gamma_\alpha(\mathbf{\Delta},\mathbf{p}_v)\right)- \mathbf{l}_{1:3}^T(\bm v+mg\mathbf{e}_3) \\
    \dot{\beta} & =  -c_2\left(\beta+\gamma_\beta(\mathbf{\Delta},\mathbf{p}_v)\right)- \mathbf{l}_{4:6}^T(\bm v+mg\mathbf{e}_3).
\end{align}  
\end{subequations}

Following the design of the nonlinear disturbance observer, the closed-loop dynamics of the quadrotor under the new DOBC scheme \eqref{eqn:controlLaw} can be summarised in the following Theorem.  

\begin{theorem}
Suppose that Assumptions \ref{sensor} and \ref{consdis} hold. Then, the system \eqref{drone} with the virtual control input \eqref{defnu} and the composite control law \eqref{eqn:controlLaw} can exponentially track the reference $\mathbf{p}_r$ over the operating range of interest, given the disturbance reconstruction \eqref{estFd} and the observer design \eqref{DOestimate}-\eqref{DOdyn}.
\end{theorem}  

\begin{IEEEproof}
The full closed-loop dynamics is analysed as follows. Let $\mathbf{e}_{Kd}=\begin{bmatrix}
    \hat K-K &\hat d-d
\end{bmatrix}^T$ be the estimation error vector. By \eqref{estdyn}, the estimation dynamics can be written as $
  \dot {\mathbf{e}}_{Kd}=\mathbf{A}_o \mathbf{e}_{Kd}$.
Concatenating with the position tracking error $\mathbf{e}_p$ and omitting some standard derivations gives the following error dynamics for the closed-loop system:
\begin{equation}\label{eqn:closedStability}
    \begin{bmatrix}
        \dot{\mathbf{e}}_p\\
        \ddot{\mathbf{e}}_p\\
      \hdashline
        \dot {\mathbf{e}}_{Kd}
    \end{bmatrix}=  \begin{bmatrix}
         \mathbf{A}_c\bigotimes  \mathbf{I}_{3\times3}&  \mathbf{\Lambda} \\
        \mathbf{0}_{2\times6} &\mathbf{A}_o
    \end{bmatrix}\begin{bmatrix}
        {\mathbf{e}}_p\\
        \dot{\mathbf{e}}_p\\
            \hdashline
        {\mathbf{e}}_{Kd}
    \end{bmatrix}
\end{equation}
where $\mathbf{\Lambda}\in \mathbb{R}^{6\times2}$ is a constant matrix. Due to the fact that $\mathbf{A}_c$ and $\mathbf{A}_o$ are both Hurwitz, the closed-loop system is exponentially stable. This completes the proof.
\end{IEEEproof}

\begin{rem} 
Since the disturbance $K$ affects all translational states, all three sets of measurements can be used to construct its estimate. This differs from the traditional DOB formulation since the number of inputs is greater than the number of disturbances. This redundancy gives the Redundant Disturbance Observer its name.
\end{rem}
\begin{rem}\label{rem:singularity} 
The RDO can decouple the estimates of $K$ and $d$ subject to $\Delta_x^2 + \Delta_y^2\neq 0$.This is a singularity in the method, and can be seen in \eqref{eqn:gainSelection}. In the implementation, a small constant, $c_3\in\mathbb{R}_+$ is added to obtain: $1/(\Delta_x^2 + \Delta_y^2) \approx 1/(\Delta_x^2 + \Delta_y^2+c_3)$, which ensures each of $l_1,l_2,l_3,l_4$ remain finite.
\end{rem}
\begin{rem}
$\gamma_\alpha$ and  $\gamma_\beta$ in \eqref{gamma} are  Riemann–Stieltjes integral of the observer gain vectors $\mathbf{l}_{1:3}^T$ and $\mathbf{l}_{4:6}^T$ w.r.t the velocity of drone $\dot {\mathbf{p}}_v$, which is nontrivial to find an explicit solution. Here, we could use the numerical method to approximate it:
\begin{equation}
\begin{aligned}
    \gamma_\alpha(t_k) &\approx \gamma_\alpha(t_{k-1})+   m\mathbf{l}_{1:3}^T(t_{k-1}) \left(\dot {\mathbf{p}}_v (t_{k})-\dot {\mathbf{p}}_v (t_{k-1})\right)\\
   \gamma_\beta(t_k) &\approx \gamma_\beta(t_{k-1})+   m\mathbf{l}_{4:6}^T(t_{k-1}) \left(\dot {\mathbf{p}}_v (t_{k})-\dot {\mathbf{p}}_v (t_{k-1})\right)
\end{aligned}
\end{equation}
where $t_k$ is the time instant and $k\in\mathbb{N}_+$.  
\end{rem}
\begin{rem}
    This method can cope with compression and tension forces if the second case of \eqref{eqn:deltaDefinition} is removed. This might occur with an extensive spring instead of a bungee. Since this paper focuses on a bungee, the second case is included. This means the RDO can cope with taut cable when $|\mathbf{\Delta}|>0$ and reduces to a nominal model where $\mathbf{F}_d = 0 + d\mathbf{e}_3$ is a single lumped disturbance. The lack of multiple flight modes facilitates stability proofs for the observer design \eqref{gainrelation}-\eqref{eqn:gainSelection} and overall closed loop stability \eqref{eqn:closedStability}.
\end{rem}
\begin{rem}
    This method only relies on existing flight control sensors, which reduces the weight of the UAV as compared to methods which directly measure the force with a strain gauge. This also means performance is tied to the accuracy of state estimation. 
\end{rem}

\section{Performance Evaluation}\label{sec:results}

\par This section aims to verify and demonstrate the performance of the proposed DOBC design using both numerical simulation and experimental flights. Two different experiments are conducted, in the first test the UAV is tasked to follow a circular reference at a constant height, with a fixed cable location. In the second experiment, the UAV is tasked to manipulate a nonlinear mechanism mimicking the extraction of a wedged object. For this experiment, we present a comprehensive demonstration of the repeatability of the combined experimental setup and controller. The proposed disturbance observer will be compared against a standard reduced-order disturbance observer (DOB) \cite{chenDisturbanceObserverBasedControlRelated2016} and an Extended State Observer (ESO) with two augmented states for each translation direction.

The DOB is designed first, and will attempt to directly estimate the lumped disturbance $\mathbf{F}_d$. Under the mild assumption that the lumped disturbance is slowly varying, the benchmark disturbance observer can be formulated as follows. First, let the error dynamics be defined as:

\begin{equation}
    \bm{\dot{\hat{F}}_d}-\bm{\dot{F}}_d = -\bm{l_d}(\bm{\hat{F}_d} - \bm{F_d})
\end{equation}
where $\mathbf{l}_d\in\mathbb{R}^{3\times3}_+$ is a positive, diagonal, constant gain matrix. Invoking the assumption that $\bm{F_d} \approx 0$, the observer can be formulated as:
\begin{align}
    \mathbf{\dot{z}}_d &= -\mathbf{l}_d\mathbf{z}_d - \mathbf{l}_d\left(m\mathbf{\dot{p}}_v + mg\mathbf{e}_3 + \bm{\nu}\right)\\
    \mathbf{\hat{F}_d} &= \mathbf{z}_d + m\mathbf{l}_d \mathbf{\dot{p}}_v.    
\end{align}
Which is the standard reduced order form available in \cite{chenDisturbanceObserverBasedControlRelated2016}. The symbol $\mathbf{{z}}_d\in\mathbb{R}^{3\times1}$ is used for the internal variable of the observer. 
\par For the ESO, each translation channel is augmented with two states (12 in total) to estimate a disturbance with a constant rate of change. Using the input transformation \eqref{defnu}, retaining Assumption \ref{assu:Priors}, and making a second assumption that $\bm{\ddot{F}}_d\approx 0$, the ESO can take the form of:
\begin{align}
    \begin{bmatrix}
        \dot{z}_{e1}\\
        \vdots\\
        \dot{z}_{e12}
    \end{bmatrix}&=\underbrace{\begin{bmatrix}
        \mathbf{0}_{9\times3} & \mathbf{I}_9 \\ \bm{0}_{3\times3} & \mathbf{0}_{3\times9}
    \end{bmatrix}}_{\mathbf{A}_e}\begin{bmatrix} z_{e1}\\ \vdots \\z_{e12}\\\end{bmatrix} + \begin{bmatrix} \textbf{0}_{3\times3} \\ \mathbf{I}_3/m \\ \bm{0}_{3\times3} \\ \bm{0}_{3\times3}\end{bmatrix}\bm{\nu} + \mathbf{L}_e(\mathbf{y} - \mathbf{C}\mathbf{z}_e)\\
    \mathbf{y} &= \underbrace{\begin{bmatrix}\mathbf{I}_6 & 0_{6\times6}\end{bmatrix}}_{\mathbf{C}}\\
    \bm{\hat{F}_d} &= m\begin{bmatrix}
        z_{e7} & z_{e8} & z_{e9}
    \end{bmatrix}^T
\end{align}
where each of the $z_{e*}\in\mathbb{R}$ corresponds to the estimated state of the ESO, $\mathbf{L}_e\in\mathbb{R}^{12\times6}$ is the ESO gain matrix, and $\mathbf{C}\in\mathbb{R}^{6\times12}$ is the state output matrix. These two methods will be compared against the newly designed RDO. For the comparison study, the same baseline controller is used, but disturbance estimates are provided by three different observers. 

\subsection{Simulation Study}\label{sec:simResults}
\par The numerical simulation is conducted in the MathWorks Simulink environment, where the system dynamics are developed using the physical parameters for the Parrot Mambo quadcopter\footnote{The underlying quadrotor dynamics are available at https://uk.mathworks.com/help/aeroblks/quadcopter-project.html} 
with an elastic tether ($K=14$N/m, $l_0 = 0.65$m, $\mathbf{p_0} = \begin{bmatrix}
    0 & 0 & 0
\end{bmatrix}^T$) added. The quadrotor is tasked to follow a helical reference trajectory under the concurrent disturbances of $K\mathbf{\Delta}$ and a vertical force $d$.

\begin{figure}[htb]
    \centering
    \includegraphics[width = 0.4\textwidth]{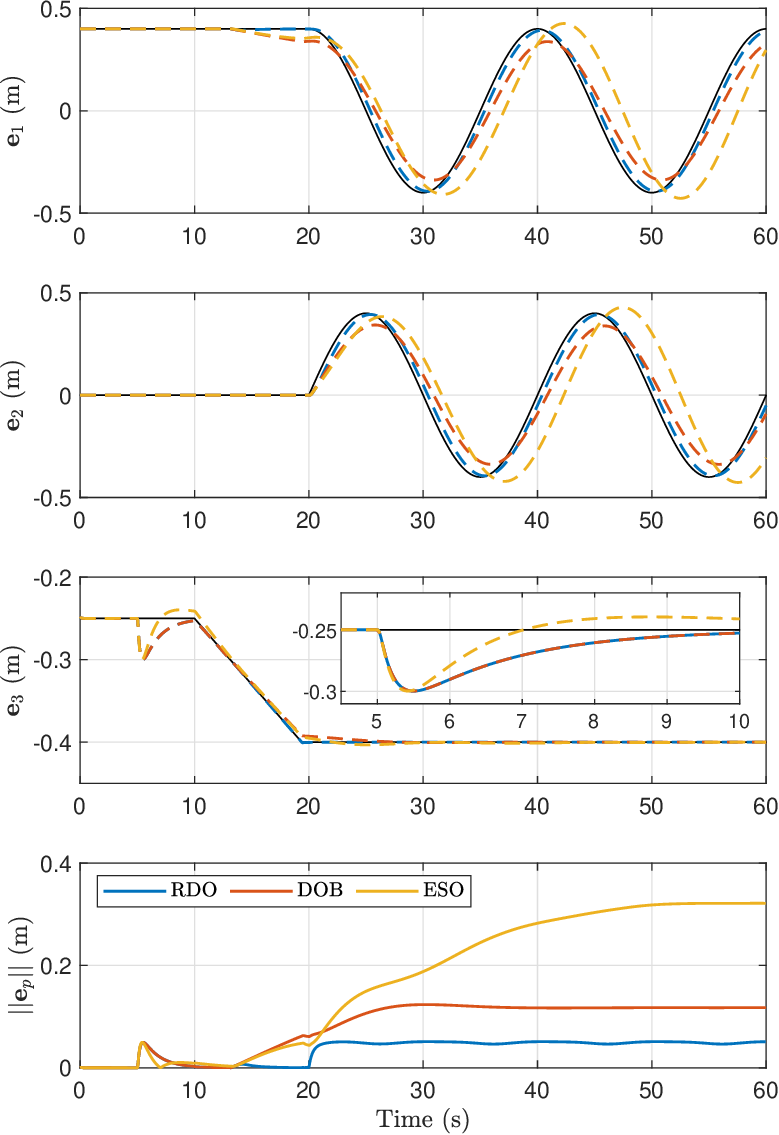}
    \caption{Tracking performance during numerical simulation with constant $||\mathbf{\Delta}||$.}
    \label{fig:simTracking}
\end{figure}

\begin{figure}[h]
    \centering
    \includegraphics[width = 0.4\textwidth]{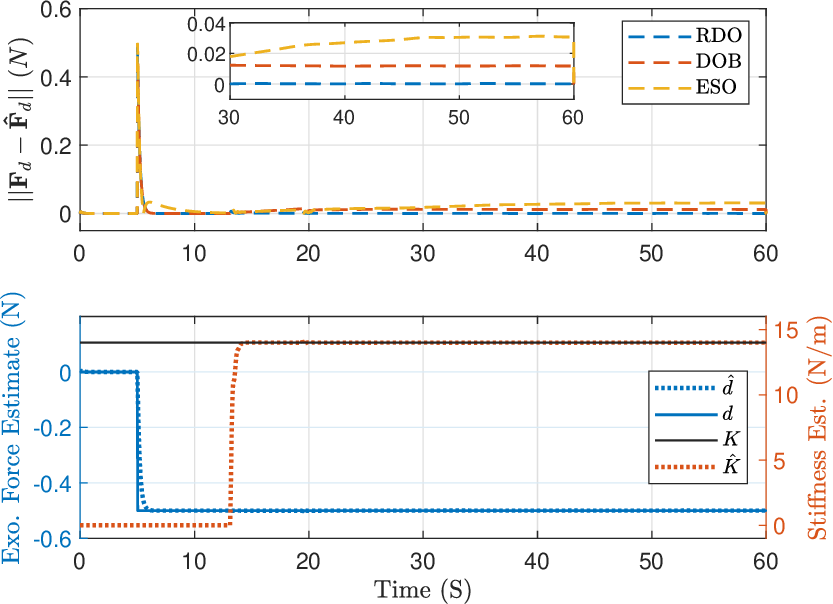}
    \caption{Estimation error of $\mathbf{F}_d$ for both observers is shown in the upper subplot. Estimation of $K$ and $d$ separately is shown in the lower subplot for the RDO.}
    \label{fig:simEstPerf}
\end{figure}

The controller settings used in the simulation are $k_p = 0.06,k_d=0.018$. To keep the test fair, the eigenvalues of each observer are set equivelantly i.e. $\bm{A_0} = \text{diag}\{-5,-5\}$ and $\mathbf{l}_d=\text{diag}\{-5,-5,-5\}$. For the ESO, the poles of the estimator are set as $\mathbf{A}_e-\mathbf{L}_e\mathbf{C} = -3\times\begin{bmatrix} 0.01\times\bm{1}_{3\times1}& 0.1\times\bm{1}_{3\times1}& 1\times\bm{1}_{3\times1}& 10\times\bm{1}_{3\times1} \end{bmatrix}^T$, where $\bm{1}_{3\times1} = \begin{bmatrix}
    1 & 1 & 1
\end{bmatrix}^T$. The choice of these poles gives an equivalent maximum error under a step disturbance for the ESO and DOB based methods. The gain $\bm{L_e}$ is then calculated with the \verb|place()| function of MATLAB. The disturbance is applied at $t=5$ seconds (see Fig. \ref{fig:simEstPerf}, lower subplot), and yields a peak overshoot of $||\bm{e}_p||\approx 0.3$m for all three methods. This can be seen in the enlarged subplot of Fig. \ref{fig:simTracking}. After the disturbance $d$ is applied, the drone is tasked to ascend to $0.4$m altitude which causes the cable to become taut at $t\approx 12$s. 

At the same time, the drone is tasked to track a circular trajectory in the $\bm{e}_1$-$\bm{e}_2$ plane. This means the disturbance in the $\bm{e}_3$ direction is constant, while the disturbance in the horizontal plane changes as the drone tracks the reference trajectory.  

Once the cable becomes taut, the RDO correctly estimates $K$ and recovers the nominal tracking performance of the baseline controller, this is shown directly in Fig. \ref{fig:simEstPerf} and discussed later. After $t=20$s, the reference reaches its maximum altitude. At this point, $||\dot{\bm{\Delta}}|| = 0$, if the controller is able to fully reject the influence of the cable. In turn, this means a successful controller will also experience $||\dot{\bm{F_d}}|| = 0$. In comparison to the proposed RDO, the DOB is able to track the reference in $\bm{e}_3$ only, since $\dot{\bm{p_r}}(t)\bm{e}_3 = 0$, which upholds the assumption in the standard DOB design. By comparison, in the horizontal plane the assumption does not hold, which damages the ability of the DOB to hold the reference. In practice, this causes a consistent tracking error of $||\bm{e}_p||\approx 0.15$m. For the ESO, the small overestimate in $\bm{\hat{\dot{F}}}_d$ causes an overshoot in the closed-loop tracking. Since the quadrotor is now further from the origin, the quadrotor now experiences a larger total disturbance - this exacerbates the error and causes both undershoot and overshoot throughout different parts of the test.

The estimation performance of the three methods are shown in Fig. \ref{fig:simEstPerf}; quantitative performance of each method is shown in table \ref{tab:objectTableResults}. In the upper subplot, the norm of the force estimation error is shown. Since the RDO can distinguish the influence of $K$ and $d$, the lower subplot shows the estimates of $\hat{K}$ and $\hat{d}$ separately. The benefit of the RDO is clearly demonstrated by this test, since the estimate $\bm{\hat{F}}_d$ is updated by $\bm{\Delta}$, the estimate can quickly be updated for any new vehicle location, which shows improved tracking performance even when compared against a method which can estimate $\bm{\dot{F}}_d$. This will be demonstrated again in the object extraction experimental case.

\begin{table}[htb]
    \centering
\begin{threeparttable}
    \caption{Quantitative performance of each controller in simulations} \label{tab:objectTableResults}
\begin{tabular}{c| c c c |p{12mm}| c}
   \hline \hline Controller & \multicolumn{3}{c|}{Pos. ISE ($m^2s$)} & \multirow{2}{12mm}{Max. Pos. Err ($m$)} & Force ISE  ($N^2s$) \\ \cline{2-4}
   & $\bm{e}_1$ & $\bm{e}_2$ & $\bm{e}_3$ & & $||\bm{F}_d-\bm{\hat{F}}_d||$ \\ \hline

   RDO & 0.05 & 0.05 & 0.003 &0.210& 0.0216 \\
   DOB & 0.28 & 0.25 & 0.003 &0.292& 0.0277\\
   ESO & 1.43 & 1.30 & 0.002 &0.569& 0.0537 \\ 
    \hline \hline
\end{tabular}
\begin{tablenotes}
\item [1]This Table shows Integral Square Error (ISE) for position tracking and force estimation, along with the maximum position error.
\end{tablenotes}
\end{threeparttable}
\end{table}

\subsection{Experimental Trajectory Tracking}\label{sec:ExpTest}
\par
\begin{figure}[htb]
    \centering
    \includegraphics[width=0.4\textwidth]{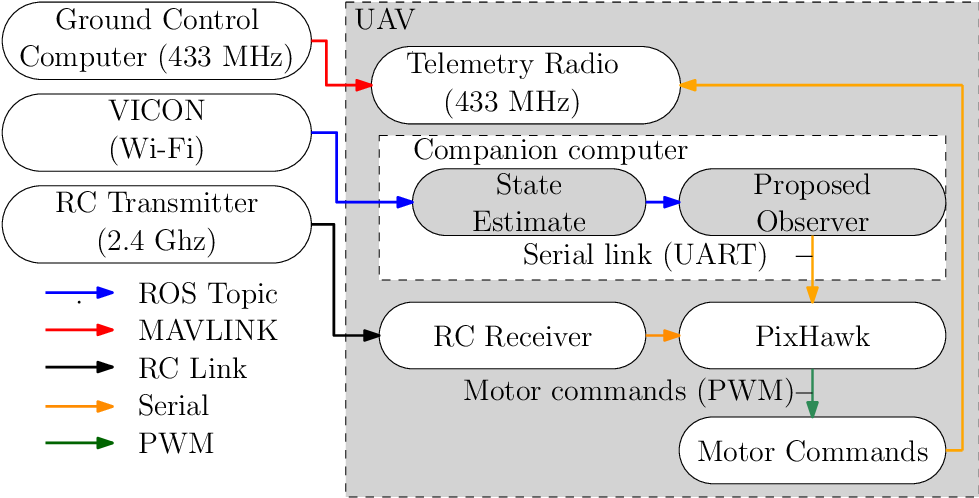}
    \caption{Hardware systems diagram. The tracking controller \eqref{eqn:controlLaw} and observer (see Sec. \ref{sec:newdesign}) are compiled to a C++ ROS node.}
    \label{fig:systemsDiag}
\end{figure}

 Following the numerical simulation, both the proposed RDO and the standard methods are deployed on the quadrotor hardware and fully tested in the flight experiments. A HexSoon-450 quadrotor ($m=1.89$kg) is used as test platform, which is affixed by a tether ($K \approx 16.5$ N/m, $l_0 = 1.4$m) to a heavy test stand. The vehicle is equipped with a PixHawk autopilot using the PX4 software and a Intel NUC onboard computer using Ubuntu 18.04 and ROS-1 Melodic. The NUC uses ROS to interact with an external VICON motion capture system to provide state measurements and run the developed algorithms both at 100 Hz. The algorithms are developed in MathWorks Simulink and compiled as a C++ ROS node. The control actions \eqref{defnu} are resolved to a thrust-attitude command \eqref{eqn:thrust-att-commands}, which are sent to the PixHawk via serial connection using an FTDI adaptor and the MAVROS protocol. As the PixHawk takes a throttle signal (0-1), the thrust command is transformed to a throttle signal by a 2D lookup table using inputs of commanded thrust and battery voltage. The system diagram is provided in Fig. \ref{fig:systemsDiag} and the test scenario is shown in Fig. \ref{fig:experimentSetup}. 

As the vehicle used in testing is significantly larger than the simulated drone, the controllers must be re-tuned. A table of tuning parameters is shown in table \ref{tab:realParams}. Controller gains were manually incremented until free-flight performance was compromised. This is the practical limit based on the available sensor and state-estimation noise. 
\begin{table}[htb]
    \centering
    \caption{Controller parameters used for hardware experiments.}
    \label{tab:realParams}
    \begin{tabular}{c | c  | c  |c }\hline \hline
        Parameter & Value & Parameter & Value  \\\hline
        Mass & 1.89 kg & $\text{eig}(\bm{A}_e-\bm{L}_e\bm{C})$ & $\begin{bmatrix}
            0.05.\bm{1}_{3\times1} \\ 0.5.\bm{1}_{3\times1} \\ 5.\bm{1}_{3\times1} \\ 25.\bm{1}_{3\times1}
        \end{bmatrix}$\\
        $\bm{l}_d$ & $\text{diag}\{0.75\,0.75\,0.75\}$ & $c_1$ & 2\\
        $k_p$ & 2.5 & $c_2$ & 0.75\\
        $k_d$ & 5 & $c_3$ & $5\times10^{-3}$\\\hline \hline
    \end{tabular}
\end{table}

\par
\begin{figure}[htb]
    \centering
    \includegraphics[width = 0.4\textwidth]{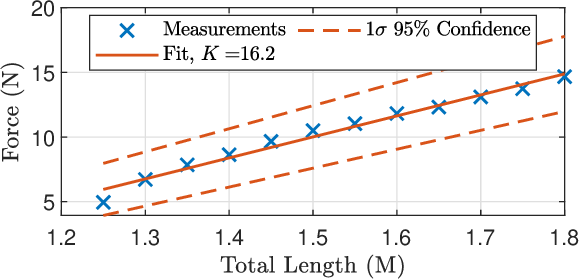}
    \caption{Data used to find the true value $K$. Including 95\% confidence bounds.}
    \label{fig:bungeedata}
\end{figure}

 To evaluate estimation accuracy, the true stiffness is measured by stretching the bungee to different lengths and recording the corresponding force. The data are displayed in Fig. \ref{fig:bungeedata}, along with the $1\sigma$ confidence bounds of the obtained stiffness.

\par In experimental testing, two scenarios were explored. First, the quadrotor must fly in a large circle with a high frequency, similar to the simulation case. This shows how the RDO can rapidly reconstruct its disturbance estimate through \eqref{eqn:deltaDefinition} and \eqref{estFd}. In the second scenario, the UAV is tasked to manipulate an overcentre mechanism. This mechanism, discussed in our previous paper \cite{marshallNovelDisturbanceDevice2023} only moves once a threshold force has been applied to the centre joint. In this work, the mechanism has been modified to release the cable as the links are actuated. The effect of this is to mimic the disturbance in a wedged object extraction case, which is similar to salient aerial manipulation tasks involving suddenly changing disturbance forces e.g., \cite{byunHybridControllerEnhancing2023,benziAdaptiveTankbasedControl2022a,cuniatoPowerBasedSafetyLayer2022}. 

\par
\begin{figure}[h]
    \centering
    \includegraphics[width = 0.4\textwidth]{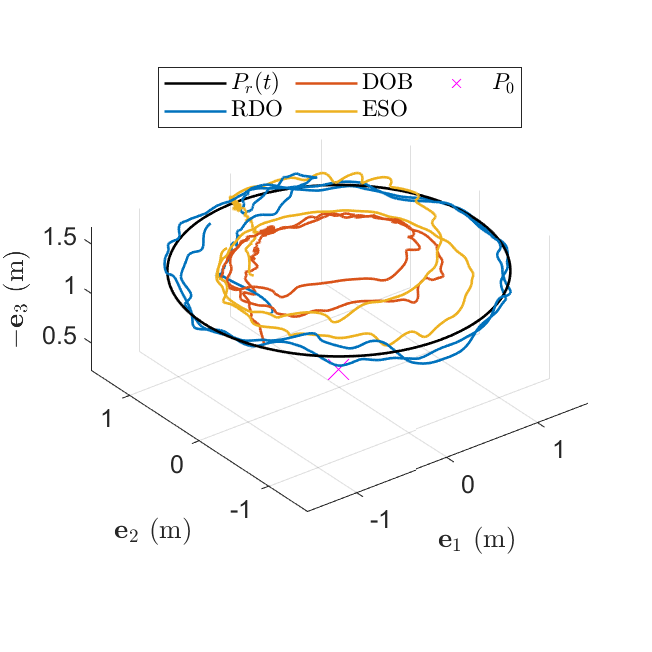}
    \caption{Isometric view of experimental test with a circular trajectory.}
    \label{fig:arciso}
\end{figure}

\begin{figure}[htb]
    \centering
    \includegraphics[width = 0.38\textwidth]{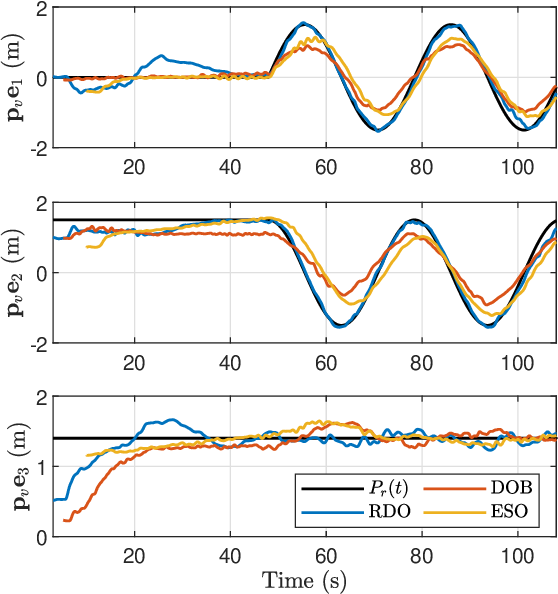}
    \caption{Time history for circular flight test}
    \label{fig:arctime}
\end{figure}

 The test data for the circular trajectory is shown as an isometric view in Fig. \ref{fig:arciso}, and a time history in Fig. \ref{fig:arctime}. The reference has a period of 30 seconds and radius of 1.5m, at a constant altitude of 1.25m. This means the magnitude of the disturbance is approximately constant as cable is stretched to a fixed length, but the disturbance components along $\bm{e}_1$ and $\bm{e}_2$ directions vary as the quadrotor follows the flight path.

\par
\begin{figure}[htb]
    \centering
    \includegraphics[width = 0.4\textwidth]{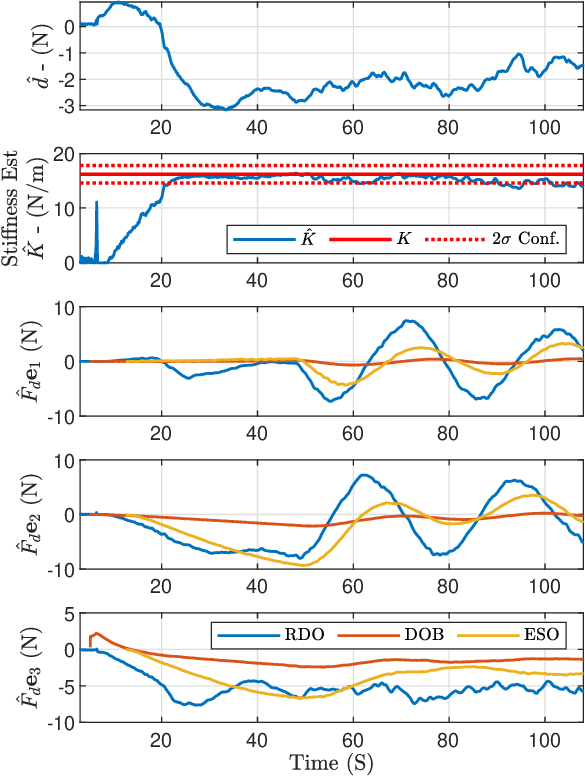}
    \caption{Stiffness estimation and force estimation in circular flight.}
    \label{fig:CircularEstimation}
\end{figure}

 Similar to the simulation study, in the circular flight test, the standard DOB design was unable to track the reference. The observer gives a steady error under this type of disturbance. The ESO presents improved tracking performance over the DOB in the experimental setup, however is still unable to fully track the reference. By comparison, the new approach precisely estimates the cable stiffness (see Fig.\,\ref{fig:CircularEstimation}) which considerably improves tracking performance as the force estimate can be quickly updated. The numerical performance of the three methods is shown in table\,\ref{tab:simtableresults}. 

\begin{table}[htb]
    \centering
    \caption{Quantitative results for setpoint-tracking experiment.}
    \label{tab:simtableresults}
\begin{tabular}{c |c c c| c }
   \hline \hline  Controller & \multicolumn{3}{c |}{Pos. ISE ($m^2s$)} & Max. Err. \\ \cline{2-4}
 & $\bm{e}_1$ & $\bm{e}_2$ & $\bm{e}_3$ & $||\bm{e}_p||$
   
   (m) \\ \hline
    RDO  & 0.40 & 0.27 & 0.27 & 0.22   \\
    DOB  & 8.57 & 15.44 & 0.65 & 1.06   \\
    ESO  & 7.79 & 13.96 & 0.63 & 1.13   \\
    \hline \hline
\end{tabular}
\end{table}

\par In Fig. \ref{fig:CircularEstimation}, the estimation performance of the RDO alongside the control signals is displayed. Here, it can be seen that the stiffness $K$ is correctly estimated to within the $1\sigma$ confidence intervals from the characterisation data. Additionally, it is clear that the RDO can very quickly update its control signals during the circular test, as the force estimate $\mathbf{\hat{F}}_d$ is updated by the state $\mathbf{\Delta}$ \eqref{eqn:deltaDefinition}. By comparison, the standard approach struggles to track the changing disturbance due to the assumption $\mathbf{\dot{F}}_d\approx 0$. 

As the tracking performance is compromised, the standard approach does not stretch the cable as much as the RDO. This results in it experiencing smaller forces, which explains the small control actions seen in the lower half of Fig. \ref{fig:CircularEstimation}. 

\subsection{Object Extraction and Repeatability}\label{sec:objext}
\par To mimic the disturbance caused by extracting an object wedged into the environment, the UAV is attached to an overcentre mechanism. This nonlinear mechanism resists force until critical load is applied. At this load the mechanism will suddenly move and release the cable. Further details on the use of the mechanism for aerial manipulation experiments can be found in our earlier work \cite{marshallNovelDisturbanceDevice2023}. Compared to an electronic release, this more closely compares to real experiments as the mechanism creates a longer and more nonlinear release from its initial position. For this test, the same hardware set-up is used, as detailed in Sec. \ref{sec:ExpTest}. As a physical mechanism is used, it is important to show the repeatability of the experiment. To this end, 9 flights are conducted for each method. To reflect this, graphical results are shown with a mean at each time sample, and numerical results are presented with their standard deviations. 

\par
\begin{figure}[htb] 
    \centering
  \subfloat[Initial position, cable slack.]{%
       \includegraphics[width=0.45\linewidth]{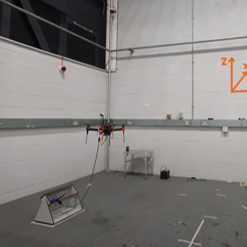}}
    \hfill
  \subfloat[Tracking ramp signal, disturbance increases.]{%
        \includegraphics[width=0.45\linewidth]{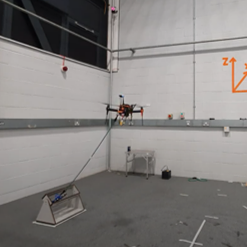}}
    \\
  \subfloat[Force threshold is exceeded and cable is released.]{%
        \includegraphics[width=0.45\linewidth]{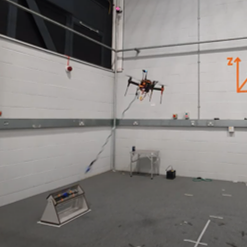}}
    \hfill
  \subfloat[Recover to nominal flight.]{%
        \includegraphics[width=0.45\linewidth]{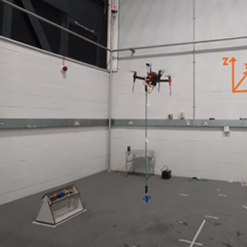}}
  \caption{Key steps in overcentre manipulation tests. For more details on the use of an overcentre mechanism for UAM testing, see earlier work in \cite{marshallNovelDisturbanceDevice2023}.}
  \label{fig:overcentreSteps} 
\end{figure}

\begin{figure}[htb]
    \centering
    \includegraphics[width = 0.38\textwidth]{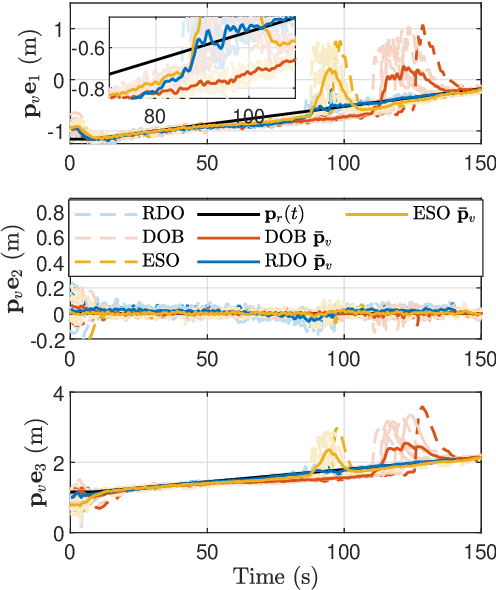} 
    \caption{Tracking performance for each method in object extraction repeatability tests. The flight with the worst maximum error magnitude is shown with a heavier dashed line.}
    \label{fig:overcentreCL}
\end{figure}

 To perform the task, the UAV begins attached to the overcentre mechanism with the cable slack. The mechanism is inclined at 45 degrees about the $\mathbf{e}_2$ direction and the UAV is tasked to track a ramp signal in the $\mathbf{e}_1$ and $\mathbf{e}_3$ directions. As the cable is stretched, the UAV will attempt to recover tracking performance under the influence of the cable. Once the cable force has reached the critical threshold for the mechanism, the links move, which unhooks the vehicle from the mechanism, resulting in the disturbance force suddenly disappearing. In Fig. \ref{fig:overcentreSteps} the experimental set-up for this test is shown.
To show the repeatability of the experiment, 9 flight tests are conducted for each observer, pale trajectories show single flights and a solid line shows the mean of all flights at each time sample. The tracking performance of each method is shown in Fig. \ref{fig:overcentreCL}. The worst-performing flight for each method uses a darker line that the other trajectories used in the data-set. Use this result for an example of a single test, and consider the mean-line as a graphical display for the repeatability for each controller.
\par
\begin{figure}[htb]
    \centering
    \includegraphics[width = 0.375 \textwidth]{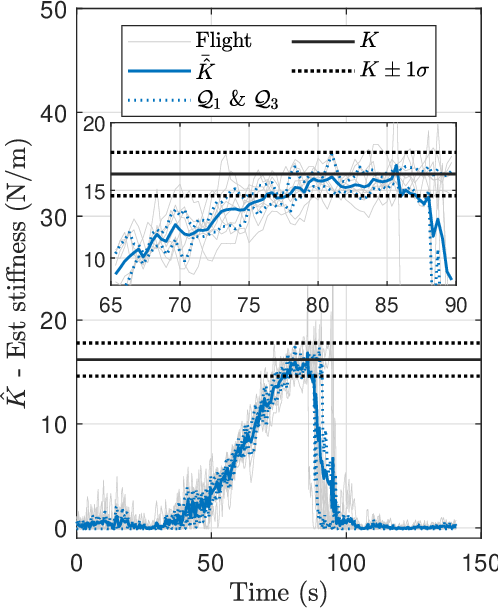}
    \caption{RDO Stiffness estimation during object extraction experiment.}
    \label{fig:ocKEst}
\end{figure}

\begin{figure}
    \centering
    \includegraphics[width = .375 \textwidth]{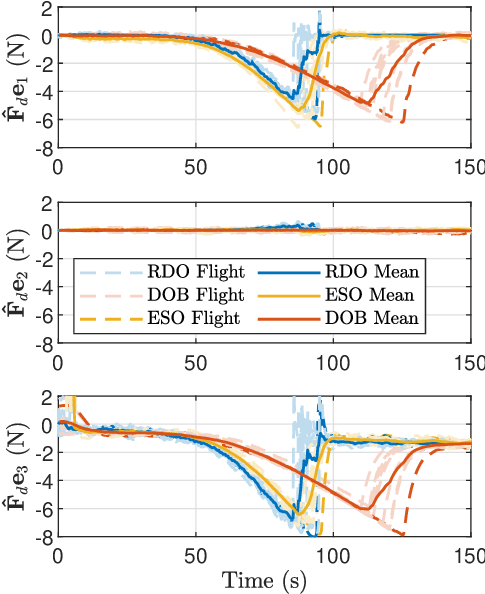}
    \caption{Control signals during object-extraction testing. The flight with the worst maximum error magnitude is shown with a heavier dashed line.}
    \label{fig:obsError}
\end{figure}

 Throughout each test, the RDO is first able to recover from a small tracking error to the reference. This is visible in the top subplot of Fig. \ref{fig:overcentreCL}. Notice as well, that the ESO is capable of tracking the reference similarly in this part of the test. This is because the spring part of the disturbance $K\bm{\Delta}$ has an nearly constant derivative when the reference signal is a ramp. In this part of the flight, the DOB lags behind the other two methods. Once the force threshold is exceeded, the tether is free to move under the bungee force. This very quickly moves the anchor point $\bm{p}_0$ underneath the drone, where it briefly oscillates.
At the point of release, the RDO displays a negligible level of overshoot. This is due to the combined effect of updating  $\bm{\hat{F}}_d$ through the state $\bm{\Delta}$ and the reduction in $\hat{K}$ (see Fig.\,\ref{fig:ocKEst}), which quickly reduces the control action. This is shown in Fig.\,\ref{fig:obsError}. Once the cable has detached from the mechanism and comes to rest below the UAV, the force from the cable disappears and reduces to the weight of the cable and hook. This motivates the RDO estimate $\hat{K}\rightarrow 0$ after the mechanism actuation. Quantitative results are displayed in table \ref{tab:qresults}. Here, the mean Integral Squared Error (ISE) over 9 flights is shown for each observer. The maximum norm of the error for the worst flight is also shown, along with the standard deviations for each value. 
\begin{table}[htb]
    \centering
\begin{threeparttable}
    \caption{Quantitative results for object-extraction task} \label{tab:qresults}
    \begin{tabular}{p{7mm}  |p{4mm} p{4mm} p{5mm} | p{4mm} p{4mm} p{4mm}  | p{7mm}| p{8.5mm}}
    \hline \hline
        Con. & \multicolumn{3}{c|}{mean ISE ($m^2s$)} & \multicolumn{3}{c|}{std. ISE $\sigma$}  &  Max. Error & std. $\sigma$  \\ \cline{2-7}
        &$\mathbf{e}_1$&$\mathbf{e}_2$&$\mathbf{e}_3$&$\mathbf{e}_1$&$\mathbf{e}_2$&$\mathbf{e}_3$&$||\mathbf{e}_p||$ &$\sigma(||\mathbf{e}_p||)$\\
        \hline
        RDO & 2.75 & 0.49 & 0.56 & 0.23 & 0.12 & 0.12 & 0.45 & 0.07 \\
        DOB & 29.86 & 0.21 & 32.69 & 5.23 & 0.03 & 6.88 & 2.14 & 0.19 \\ 
        HESO & 18.72 & 0.19 & 11.69 & 1.67 & 0.05 & 1.70 & 1.77 & 0.08 \\
        \hline \hline
    \end{tabular}
\begin{tablenotes}
\item [1] The results include the Integral Square Error and maximum error. ISE is computed as a mean over 9 flights, maximum error is calculated from the worst-case result of each 9 flights.
\end{tablenotes}
\end{threeparttable}
\end{table}

\section{Conclusions}\label{sec:concs}
\par This paper presents a redundant disturbance observer based approach to estimate cable stiffness and additional vertical disturbances for a tethered quadcopter. By modelling the cable as a spring, only two constant parameters need to be estimated, namely the stiffness $K$ and a vertical disturbance $d$, allowing the state-dependent disturbance to be reconstructed. To exploit redundant measurements for this estimation task, the observer gains are allocated via a pseudo-inverse process, ensuring the estimates are stable and decoupled. With the proposed design, the tracking performance is significantly improved over standard DOBC and ESO approaches. This is tested in a trajectory tracking task, and with the actuation of a nonlinear mechanism which mimics the forces involved in an object-extraction task. The experimental results also present the characterisation of the real cable and show that the estimated value lies within the $1\sigma$ confidence bounds of the characterised value. The new method does this without requiring any additional force sensors provided that the anchor location and tether length are known. By treating the stiffness as an unknown estimation parameter, this approach also simplifies control design, since multiple flight modes are not required for attached and detached cases, which may lead to new applications of aerial robotics to interact with the environment via elastic and flexible connections.

\bibliographystyle{IEEEtran}
\bibliography{MyLibrary}

\end{document}